\documentclass[11pt]{article}
\usepackage[preprint]{acl}
\usepackage{times}
\usepackage{latexsym}
\usepackage[T1]{fontenc}
\usepackage[utf8]{inputenc}
\usepackage{microtype}
\usepackage{inconsolata}
\usepackage{graphicx}
\usepackage{float}
\usepackage{booktabs}
\usepackage{multirow}
\usepackage{subcaption}

\title{Fewer Heads, Fewer Memories: Pruning Redundant Attention Heads Suppresses Verbatim Memorisation} 

\author{Aryan Chowdhury\\
  Visiting Researcher \\ University of Manchester \\
  \\\And
  Riza Batista-Navarro \\
  Senior Lecturer in Text Mining \\ University of Manchester \\
  }

\begin{document}
\maketitle

\begin{abstract}
Large language models have been shown to memorise and reproduce verbatim sequences from their training data, raising concerns about privacy leakage, intellectual property exposure, and the limits of generalisation. Recent theoretical work suggests that attention heads may contribute to this behaviour by functioning as associative memory mechanisms, with memorisation capacity scaling with the number of heads. At the same time, empirical work has shown that many attention heads are redundant and can be pruned with limited impact on downstream performance. Despite these parallel findings, the effect of pruning redundant heads on memorised-sequence reproduction remains unexplored. In this work, we demonstrate that zeroing out a substantial fraction of attention heads prior to inference can significantly suppress verbatim memorisation while preserving, or even improving, language modelling performance on the benchmark used to identify redundant heads. Evaluating GPT-2-XL(1.5B) on known memorised sequences, we find that pruning 729 task-redundant heads reduces the memorisation count by 91.5\%, while improving WikiText-103 perplexity by 0.92. These findings suggest that attention-head redundancy can be exploited to mitigate memorisation at inference time, without retraining or modifying the training data.
\end{abstract}

\section{Introduction}

Large language models have been shown to memorise and reproduce verbatim sequences from their training data, raising concerns about privacy leakage, intellectual property exposure, and the limits of generalisation \citep{carlini2021extracting, carlini2022quantifying}. Existing mitigations, such as deduplication, differential privacy, and unlearning \citep{kandpal2022deduplicating, yu2021differentially, si2023knowledge}, either intervene at training time or require identifying specific memorised content, making them difficult to apply to deployed models.

Prior work has shown that MLP layers store substantial factual and memorised knowledge \cite{NEURIPS2022_6f1d43d5, nanda2023factfinding}, while the role of attention remains comparatively underexplored. Recent theoretical results establish that multi-head attention can function as an associative memory, with memorisation capacity scaling in the number of heads \citep{mahdavi2024memorization, dana2024memorization}. This capacity arises from softmax saturation: when attention weights concentrate sharply on a small set of positions, the corresponding value vectors dominate the head's output, enabling precise token retrieval rather than diffuse contextual aggregation. Independently, empirical work has consistently shown that a substantial fraction of attention heads can be zeroed out at inference time without degrading downstream task performance \citep{michel2019sixteen, voita2019analyzing}. The implications of this redundancy for memorisation, however, remain unexplored. Our contributions are as follows:
\begin{itemize}
    \item We demonstrate that inference-time pruning of task-redundant attention heads substantially reduces memorised-sequence reproduction while preserving language modelling performance on the benchmark used to identify redundant heads.
    \item We find that memorised sequences induce sharper softmax saturation and stronger cross-head agreement, creating a coordinated signal toward the memorised continuation.
    \item We show that pruning task-redundant heads weakens this collective signal by removing part of their contribution from the residual stream, offering a mechanistic explanation for the observed reduction in memorisation.
\end{itemize}

\section{Methodology}

\subsection{Verbatim Memorisation}
Following prior work, we define memorisation using $k$-extractability. A sequence $s$ is considered memorised if, when prompted with its first $k$ tokens, the model reproduces the next $k$ tokens verbatim under greedy decoding \citep{biderman2023emergent}. We select $k=16$ as the shortest prefix length to capture short-span memorisation, while $k=32$ and $k=50$ follow the thresholds used in prior extractability studies \citep{carlini2021extracting, biderman2023emergent}.

\subsection{Models and Datasets}

We conduct experiments on Pythia-2.8B-Deduped, Pythia-6.9B-Deduped, and GPT-2-XL (1.5B) \citep{biderman2023pythia, radford2019language}, chosen for their open weights, documented pretraining corpora, and established use in memorisation research. Together, they span a range of scales and architectures, which is relevant given that memorisation rates increase with model capacity \citep{carlini2022quantifying}.

For each model, memorised and non-memorised sequences are drawn from its corresponding pretraining corpus: Pile-Deduped for the Pythia models and OpenWebText for GPT-2-XL \citep{gao2020pile, radford2019language}. For the Pythia models, we sample 20,000 memorised sequences per model from the public list of memorised sequences \citep{biderman2023emergent}. For GPT-2-XL, we construct an equivalent set by applying the criterion from Section 2.1 to every OpenWebText sequence, yielding between 4,800 and 5,590 unique memorised sequences depending on $k$. In all cases, memorisation count is measured as the number of sequences reproduced verbatim under greedy decoding, before and after head pruning. For each model, we additionally construct a matched non-memorised set by sampling an equal number of sequences whose greedy continuation does not exactly match the target continuation.

Performance is evaluated on WikiText-103 and LAMBADA, standard benchmarks for language modelling and next-token prediction \citep{merity2016wikitext, paperno2016lambada}; we report perplexity on both and last-word prediction accuracy on LAMBADA. For gate optimisation (Section~\ref{sec:gate-training}), we use a held-out split of WikiText-103/LAMBADA, disjoint from the split used for final perplexity and accuracy reporting.

\subsection{Identifying Task-Redundant Heads}\label{sec:gate-training}

We identify task-redundant attention heads by learning sparse binary gates over head outputs while keeping all model parameters frozen. For each head \(h\) in layer \(l\), the gated output is
\[
z'_{l,h} = b_{l,h} z_{l,h},
\qquad
b_{l,h} = \mathbf{1}[\sigma(\alpha_{l,h}) > 0.5],
\]
where \(z_{l,h}\) is the head output, \(\alpha_{l,h}\) is a learned gate logit, and \(b_{l,h}\) is a binary gate. Since \(b_{l,h}\) is a non-differentiable step function of \(\alpha_{l,h}\), we use a straight-through estimator \citep{bengio2013estimating}: hard gates are used in the forward pass, while gradients are passed through the sigmoid relaxation \(\sigma(\alpha_{l,h})\) in the backward pass, allowing standard backpropagation despite the non-differentiable forward operation.
\[
\mathcal{L}
=
\max\left(0,\mathcal{L}_{\mathrm{gated}}-\mathcal{L}_{\mathrm{orig}}-\log r\right)
+
\lambda \sum_{l,h} \sigma(\alpha_{l,h}),
\]
where \(r\) is the allowed perplexity ratio and \(\lambda\) controls sparsity. Since perplexity is exponential in loss, capping the perplexity ratio at \(r\) caps the increase in loss at $\log r$. The objective thus discourages pruning configurations that exceed the permitted increase in loss, while encouraging fewer active heads. 

\paragraph{Implementation details.} Gate logits $\alpha_{l,h}$ are initialised to $5.0$ for all heads, corresponding to $\sigma(\alpha_{l,h}) \approx 0.99$, so that all heads begin active and pruning proceeds from the full model. We set $r=1.00$ across pruning objectives and $\lambda=10^{-4}$, and optimise gate logits with AdamW \citep{loshchilov2017decoupled} at a learning rate of $10^{-2}$, using batches of 8 sequences sampled with replacement from the held-out gate-optimisation split, for a fixed budget of 1,000 steps. Rather than using the gate configuration from the final training step, we track, at each step, the batch-level perplexity ratio $\mathcal{L}_{\mathrm{gated}}/\mathcal{L}_{\mathrm{orig}}$ under the current hard gates, and select the gate configuration achieving the lowest observed ratio over the full training run. Heads with inactive gates in the selected configuration are zeroed out at inference time.

\subsection{Mechanistic Analysis Metrics}

To analyse how pruning affects memorisation-related computation, we compute four diagnostics on both memorised and non-memorised sequences.
For each attention head, we measure Shannon entropy as:
\[
H_{l,h}
=
-\frac{1}{N}
\sum_{n=1}^{N}
\frac{1}{T_n}
\sum_{t=1}^{T_n}
\sum_{j \leq t}
A^{(n)}_{l,h,t,j}
\log A^{(n)}_{l,h,t,j},
\]
where \(A_{l,h,t,j}\) is the attention weight from position \(t\) to position \(j\), averaged across sequences. Lower entropy indicates more concentrated attention, while higher entropy indicates more diffuse contextual mixing.

We measure softmax saturation as the number of attention rows with maximum weight above a threshold of \(0.95\), aggregated across token positions and sequences. This captures the extent to which a head places nearly all of its attention mass on a single position while remaining permissive enough to surface meaningful differences in saturation across sequence types \citep{mahdavi2024memorization}.

To assess whether saturated heads coordinate on the same evidence, we measure cross-head agreement as the number of times attention rows from different heads share the same maximum-weight position. Higher agreement indicates that heads are converging on the same position rather than attending independently.

Finally, we measure the top-two logit margin, $\Delta_{\mathrm{logit}} = \ell_{(1)} - \ell_{(2)}$, averaged across all generated positions. A larger margin indicates a stronger model preference for a single continuation. Comparing each diagnostic before and after pruning allows us to test whether head removal weakens the collective signal driving memorised reproduction.

To support reproducibility, we will release all the source code upon acceptance. We also acknowledge the use of AI assistance during the preparation of this work for grammar and spell checking, code assistance, and improving clarity and readability.

\section{Results}

\subsection{Effects of Pruning on Memorisation}

\begin{table*}[t]
\centering
\footnotesize
\setlength{\tabcolsep}{4pt}
\begin{tabular}{llccccc}
\toprule
& & & \textbf{LM Performance} & \multicolumn{3}{c}{\textbf{Memorisation Count}} \\
\cmidrule(lr){4-4} \cmidrule(lr){5-7}
\textbf{Model} & \textbf{Variant} & \textbf{Active Heads} & \textbf{WikiText (PPL)} & \textbf{$k{=}16$} & \textbf{$k{=}32$} & \textbf{$k{=}50$} \\
\midrule
\multirow{3}{*}{GPT-2-XL(1.5B)}
 & Baseline           & 1200 / 1200 & 10.08  & 5590 & 5540 & 4800 \\
 & Random             & 471 / 1200    & 146.82 & 140  & 330  & 910 \\
 & Pruned (WT)        & 471 / 1200  & 9.16   & 315  & 473  & 1103 \\
\midrule
\multirow{3}{*}{Pythia-2.8B}
 & Baseline           & 1024 / 1024   & 7.56   & 20,000 & 20,000 & 20,000 \\
 & Random             & 628 / 1024 & 18.30  & 5365  & 4326 & 4661 \\
 & Pruned (WT)        & 628 / 1024 & 7.32   & 4726 & 2875 & 4770 \\
\midrule
\multirow{3}{*}{Pythia-6.9B}
 & Baseline           & 1024 / 1024   & 6.90   & 20,000 & 20,000 & 20,000 \\
 & Random             & 650 / 1024    & 17.20  & 4905  & 4751  & 4310 \\
 & Pruned (WT)        & 650 / 1024    & 6.73   & 4221 & 3500 & 4202 \\
\bottomrule
\end{tabular}
\caption{Language modelling performance and memorisation counts (k-extractability) under WikiText-103-guided pruning (WT), alongside baseline and random-pruning conditions.}
\label{tab:pruning-results-wt}
\end{table*}

\begin{table*}[t]
\centering
\footnotesize
\setlength{\tabcolsep}{4pt}
\begin{tabular}{llccccc}
\toprule
& & & \textbf{LM Performance} & \multicolumn{3}{c}{\textbf{Memorisation Count}} \\
\cmidrule(lr){4-4} \cmidrule(lr){5-7}
\textbf{Model} & \textbf{Variant} & \textbf{Active Heads} & \textbf{LAMBADA (PPL/Acc.)} & \textbf{$k{=}16$} & \textbf{$k{=}32$} & \textbf{$k{=}50$} \\
\midrule
\multirow{3}{*}{GPT-2-XL(1.5B)}
 & Baseline           & 1200 / 1200 & 25.97/0.6128 & 5590 & 5540 & 4800 \\
 & Random             & 642 / 1200    & 91.38/0.2014 & 140  & 330  & 910 \\
 & Pruned (LB)        & 642 / 1200    & 25.09/0.6148 & 1334 & 1133 & 2200 \\
\midrule
\multirow{3}{*}{Pythia-2.8B}
 & Baseline           & 1024 / 1024   & 17.35/0.7489 & 20,000 & 20,000 & 20,000 \\
 & Random             & 775 / 1024 & 33.24/0.3617 & 7981  & 7326 & 11661 \\
 & Pruned (LB)        & 775 / 1024 & 17.51/0.7543 & 8094 & 6995 & 10414 \\
\midrule
\multirow{3}{*}{Pythia-6.9B} 
 & Baseline           & 1024 / 1024   & 14.80/0.7812 & 20,000 & 20,000 & 20,000 \\
 & Random             & 780 / 1024    & 29.40/0.4100 & 10302  & 7503  & 11320 \\
 & Pruned (LB)        & 780 / 1024    & 14.95/0.7760 & 8500 & 7222 & 11101 \\
\bottomrule
\end{tabular}
\caption{Language modelling performance and memorisation counts (k-extractability) under LAMBADA-guided pruning (LB), alongside baseline and random-pruning conditions.}
\label{tab:pruning-results-lb}
\end{table*}

Across all models and benchmarks, Tables~\ref{tab:pruning-results-wt} and~\ref{tab:pruning-results-lb} show that pruning substantially reduces memorised-sequence reproduction while preserving, and in several cases improving, performance on the benchmark used to identify redundant heads. We first compare the Baseline and Pruned conditions, leaving Random to be discussed separately below.

Notably, the sets of heads identified as task-redundant differ between the WikiText-103 and LAMBADA objectives, and both differ across models, yet both configurations suppress memorisation substantially in every setting. This suggests that memorisation is not localised in a single narrow subset of heads, but distributed across the attention mechanism, with different task-preserving subsets still capable of suppressing it when removed.

Pruning against the WikiText split (WT) consistently improves or preserves WikiText perplexity while removing the larger share of heads and yielding the larger memorisation reduction. GPT-2-XL improves from 10.08 to 9.16 while removing $\sim$61\% of heads and cutting memorisation from 5590 to 315 at $k{=}16$, and both Pythia models show the same pattern (7.56$\to$7.32 and 6.90$\to$6.73), with $k{=}32$ memorisation falling from 20,000 to 2875 and 3500. 

Pruning against the LAMBADA split (LB) shows the complementary pattern: fewer heads are removed, and the memorisation reduction is correspondingly smaller, but LAMBADA performance is preserved or slightly improved (e.g.\ GPT-2-XL: 25.09/0.6148 vs.\ 25.97/0.6128 baseline). Together, these results indicate that memorisation reduction scales with the number of heads pruned, regardless of which benchmark's redundant heads are targeted.

This relationship does not imply that removing heads indiscriminately would suppress memorisation just as well. As the random baseline below shows, the identity of the removed heads matters as much as their number.

To isolate this, we include a random baseline in which the number of pruned heads is set to the same value as the number of heads pruned using the sparse gates. We run each model 5 times, each with a different random selection of heads, and report the average language modelling performance and memorisation counts (rounded up to the nearest integer).

Although random pruning sometimes yields lower memorisation counts than targeted pruning (e.g., 330 vs.\ 473 memorised sequences at $k{=}32$ for GPT-2-XL on WT pruning), this comes at the cost of severe degradation in language modelling capability. Under WikiText-guided pruning, GPT-2-XL's WikiText perplexity rises from 10.08 to 146.82 under random pruning, with both Pythia models showing comparable degradation (7.56$\to$18.30 and 6.90$\to$17.20). Under LAMBADA-guided pruning, GPT-2-XL's LAMBADA accuracy falls from 0.6128 to 0.2014 under random pruning, with both Pythia models again showing substantial degradation (0.7489$\to$0.3617 and 0.7812$\to$0.4100). Since a model with severely impaired predictive ability will trivially fail to reproduce training sequences, memorisation count alone is an insufficient measure of a mitigation strategy's value. In contrast, task-guided pruning achieves a substantially better utility--memorisation trade-off by reducing extractability while retaining the model's language modelling ability.

\subsection{Attention Saturation}

For the mechanistic analysis, we focus on Pythia-2.8B and GPT-2-XL pruned under the WikiText-103 objective. We exclude Pythia-6.9B due to computational resource constraints, and restrict to a single benchmark objective for clarity of exposition. 

\begin{table}[t]
\centering
\resizebox{\columnwidth}{!}{%
\begin{tabular}{llccccc}
\hline
\textbf{Model} & \textbf{Variant} & \textbf{Active Heads} & \textbf{Mem. Total} & \textbf{Non-mem. Total} & \textbf{$\Delta$ Total} & \textbf{$\Delta$ Average} \\
\hline
Pythia-2.8B & Baseline & 1024 & 2129.51 & 2265.23 & -135.72 & -0.1325 \\
Pythia-2.8B & Pruned   & 628  & 2289.60 & 2316.41 & -26.81  & -0.0427\\
GPT-2-XL    & Baseline & 1200 & 1527.66 & 1554.35 & -26.69  & -0.0222 \\
GPT-2-XL    & Pruned   & 471  & 1663.92 & 1614.55 & 49.37   & 0.1048 \\
\hline
\end{tabular}%
}
\caption{Attention entropy on memorised and non-memorised sequences, before and after pruning. Total and Average are computed over active heads only, so that values before and after pruning are directly comparable at the per-head level. $\Delta$ Total and $\Delta$ Average denote the memorised value minus the non-memorised value.}
\label{tab:entropy_normalised}
\end{table}

\begin{table}[t]
\centering
\resizebox{\columnwidth}{!}{%
\begin{tabular}{llcccccc}
\hline
\textbf{Model} & \textbf{Variant} & \textbf{Active Heads} & \textbf{Mem. Count} & \textbf{Non-mem. Count} & \textbf{$\Delta$ Count} & \textbf{Mem. Rate} & \textbf{Non-mem. Rate} \\
\hline
Pythia-2.8B & Baseline & 1024 & 22{,}937{,}568 & 16{,}373{,}444 & 6{,}564{,}124 & 22399.97 & 15989.69 \\
Pythia-2.8B & Pruned   & 628  & 9{,}498{,}653  & 7{,}778{,}552  & 1{,}720{,}101 & 15125.24 & 12386.23 \\
GPT-2-XL    & Baseline & 1200 & 40{,}614{,}224 & 23{,}021{,}552 & 17{,}592{,}672 & 33845.19 & 19184.63 \\
GPT-2-XL    & Pruned   & 471  & 11{,}099{,}121 & 8{,}400{,}622  & 2{,}698{,}499  & 23565.01 & 17835.72 \\
\hline
\end{tabular}%
}
\caption{Softmax saturation counts on memorised and non-memorised sequences, before and after WikiText-guided pruning, alongside the corresponding rate per active head ($\Delta$ Count = Mem. Count $-$ Non-mem. Count).}
\label{tab:saturation_rate}
\end{table}

Tables~\ref{tab:entropy_normalised} and~\ref{tab:saturation_rate} report attention entropy and softmax saturation normalised by the number of active heads, so values before and after pruning are directly comparable per head. In both unpruned models, memorised sequences yield lower attention entropy ($\Delta$ Average of $-0.1325$ for Pythia-2.8B, $-0.0222$ for GPT-2-XL) and more softmax saturation (ratios of $1.40$ and $1.76$) than non-memorised ones, consistent with memorised content eliciting more peaked, more confident attention.

Pruning narrows this saturation gap. The memorised-to-non-memorised ratio falls to $1.22$ (Pythia-2.8B) and $1.32$ (GPT-2-XL), and the absolute gap in counts shrinks by about $74\%$ and $85\%$. Saturation drops more on memorised sequences than on non-memorised ones (Pythia-2.8B: $-58.6\%$ vs.\ $-52.5\%$; GPT-2-XL: $-72.7\%$ vs.\ $-63.5\%$). Moreover, the reduction is not explained by head count alone: rates per active head fall by $32.5\%$ and $30.4\%$ on memorised text, versus $22.5\%$ and $7.0\%$ on non-memorised text, with the contrast most pronounced in GPT-2-XL.

The surviving heads also attend more diffusely. Mean per-head entropy on memorised text rises from about $2.08$ to $3.65$ in Pythia-2.8B and from about $1.27$ to $3.53$ in GPT-2-XL. Since our pruning objective does not use entropy, this cannot be attributed to the removal of low-entropy heads. A more plausible explanation is a ripple effect: removing heads alters the representations received by later layers, so the remaining heads attend less sharply and saturate less in general. We do not test this here and leave it to future work.

Consequently, the entropy gap attenuates rather than disappears. In Pythia-2.8B it shrinks by about $68\%$ (from $-0.1325$ to $-0.0427$) but keeps its sign, whereas in GPT-2-XL it reverses ($+0.1048$). Yet memorised sequences in the pruned GPT-2-XL still saturate more per head ($23{,}565$ vs.\ $17{,}836$) despite their higher mean entropy. Because saturation counts extreme peaks while mean entropy averages over the whole distribution, the pruned model may contain mostly diffuse heads with occasional sharp peaks on memorised content. We leave testing this hypothesis to future work.

Finally, the effects are stronger in GPT-2-XL, which is pruned considerably more aggressively than Pythia-2.8B and shows both larger relative reductions in saturation and the sign reversal in entropy. Architectural differences, such as positional encoding and the arrangement of attention and MLP blocks, are further candidate explanations, but the present results cannot separate them.

\begin{figure}[t]
\centering
\begin{minipage}{0.48\columnwidth}
    \centering
    \includegraphics[width=\linewidth]{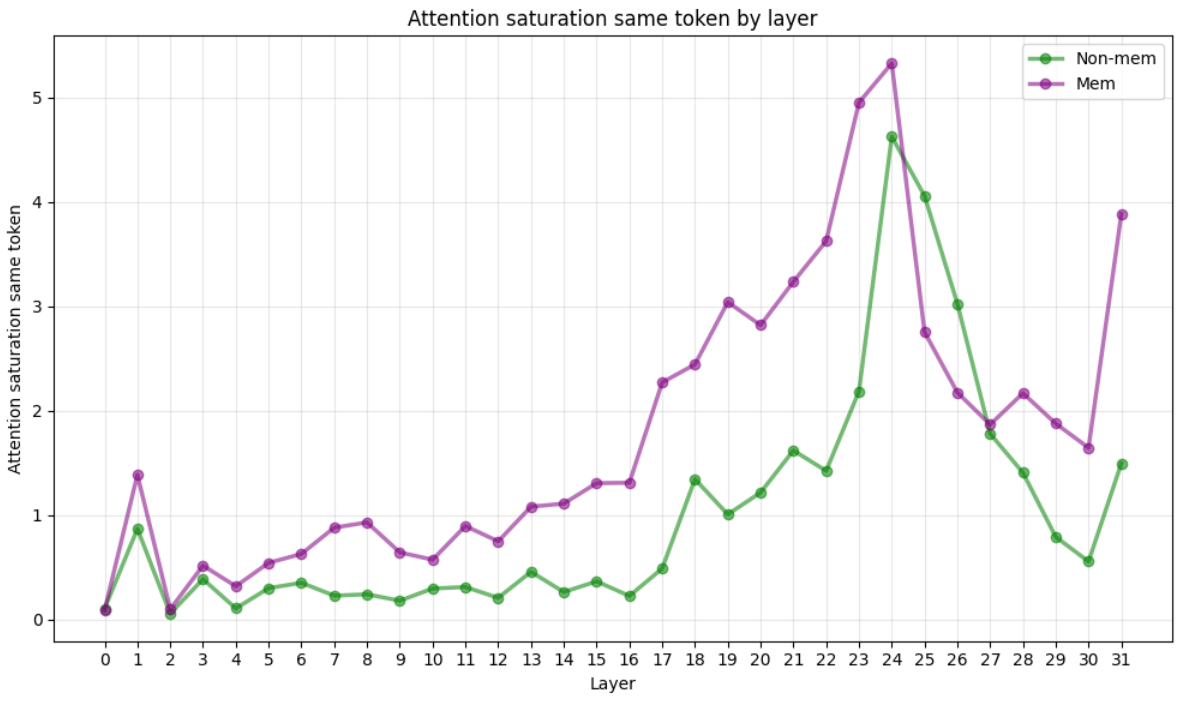}
    \subcaption{Before pruning}
    \label{fig:cross_head_agreement_baseline}
\end{minipage}
\hfill
\begin{minipage}{0.48\columnwidth}
    \centering
    \includegraphics[width=\linewidth]{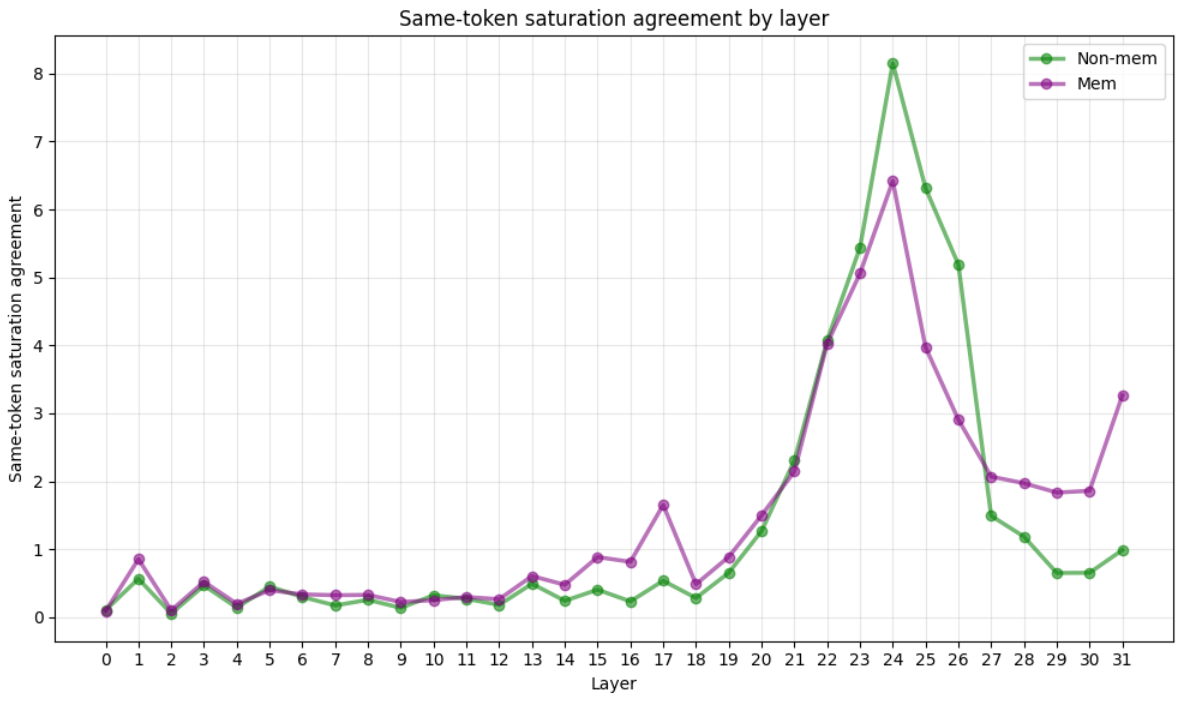}
    \subcaption{After pruning}
    \label{fig:cross_head_agreement_pruned}
\end{minipage}
\caption{Average cross-head agreement across layers among heads exceeding the softmax saturation threshold, before and after pruning.}
\label{fig:cross_head_agreement}
\end{figure}

Figure~\ref{fig:cross_head_agreement} illustrates this coordination effect before and after pruning. In the baseline model, memorised sequences elicit markedly stronger cross-head agreement among saturated heads across layers; after pruning, this gap narrows substantially, with memorised and non-memorised sequences showing more similar coordination profiles. As with the saturation count above, this comparison is drawn over the pool of heads exceeding the saturation threshold, which itself shrinks after pruning; we have not yet verified how much of the narrowed gap follows from this smaller pool versus a genuine change in coordination among the heads that remain.

\begin{table}[t]
\centering
\resizebox{\columnwidth}{!}{%
\begin{tabular}{lccc}
\hline
\textbf{Model} & \textbf{Setting} & \textbf{Non-memorised margin} & \textbf{Memorised margin} \\
\hline
Pythia-2.8B & Baseline & 2.01 & 8.79 \\
Pythia-2.8B & Pruned & 1.86 & 3.60 \\
GPT-2-XL & Baseline & 2.02 & 7.68 \\
GPT-2-XL & Pruned & 1.76 & 3.86 \\
\hline
\end{tabular}%
}
\caption{Average top-two logit margin on memorised and non-memorised sequences before and after head pruning.}
\label{tab:logit_margin}
\end{table}

The logit margin results in Table~\ref{tab:logit_margin} are consistent with this account: baseline models produce top-two margins of 8.79 and 7.68 on memorised sequences, roughly four times the non-memorised margins, indicating that the saturation and coordination patterns above co-occur with anomalously high-confidence predictions on memorised sequences. Pruning roughly halves these margins (to 3.60 and 3.86) while leaving non-memorised margins nearly unchanged, suggesting a selective effect on the mechanism rather than a general reduction in model confidence. Aggregated over 20,000 sequences, this per-step reduction in confidence is plausibly sufficient to cause the model to exit the memorised track frequently, contributing to the large count reductions in Table~\ref{tab:pruning-results-wt}. We note, as in the Conclusion, that this evidence remains correlational: it establishes that saturation, coordination, and confidence all shift together under pruning, but does not on its own establish that saturation causes the confidence effect, as opposed to both arising from a shared underlying change in the pruned heads' behaviour.

\section{Conclusion}

We have shown that inference-time attention head pruning reduces memorised sequence reproduction by 44--94\% across three models and two benchmarks, while preserving, and in several cases improving, language modelling performance on the benchmark used to identify redundant heads. The result holds across pruning objectives that identify different head subsets, indicating that memorisation is distributed across the attention mechanism rather than localised in a narrow set of critical heads. Mechanistically, memorised sequences induce more frequent and spatially coordinated softmax saturation across heads, translating into anomalously high next-token confidence that pruning selectively disrupts without degrading language modelling behaviour. However, the mechanistic account remains correlational; causal intervention analysis, a controlled comparison against non-memorised sequences matched for predictability, and probing of the MLP stream, which may retain an independent memorisation signal, are left to future work.

\section{Limitations}
This paper's focus is to demonstrate that inference-time attention-head pruning can suppress memorised-sequence reproduction, and to shed light on the mechanistic role attention saturation plays in that behaviour. Several limitations should be noted. Our memorisation evaluation uses $k$-extractability under greedy decoding, which captures only one facet of memorisation; a reduced count under this criterion is not a privacy guarantee, since verbatim reproduction is a narrow proxy \citep{ippolito2023preventing} and training data may remain recoverable under sampling-based decoding, longer prompts, or adversarial extraction. Sensitivity to decoding strategy, alongside model size, is exactly the kind of robustness question we flag as future work, since establishing it properly requires its own systematic study, and attention-head pruning itself leaves model weights unchanged, so it should be understood as an inference-time mitigation rather than unlearning. The mechanistic account, in turn, draws on theoretical results for attention-only transformers that do not directly extend to full architectures with residual connections, layer normalisation, and MLP blocks, so our saturation hypothesis is motivated by theory rather than a direct application of it. Experiments are further restricted to three decoder-only models of moderate size with documented pretraining corpora, and it is unclear whether the results generalise to larger, instruction-tuned, or architecturally distinct models; GPT-Neo's alternating local and global attention layers, for example, differ from the architectures studied here, and evaluation on such models is left to future work. Finally, task preservation is assessed on only two benchmarks, WikiText-103 and LAMBADA, which may not reflect broader capabilities such as factual recall or long-context reasoning.

\bibliography{custom}

\appendix

\end{document}